\documentclass[review]{elsarticle}

\usepackage{framed,multirow}

\usepackage[english]{babel}
\usepackage[utf8x]{inputenc}
\usepackage{gensymb}
\usepackage{amsmath}
\usepackage{graphicx}
\usepackage{epstopdf}
\usepackage{subfig}
\usepackage[colorinlistoftodos]{todonotes}
\usepackage{mathtools}
\usepackage{booktabs}
\usepackage{rotating}
\usepackage{amssymb}
\usepackage{latexsym}
\usepackage{comment}

\journal{Computer Vision and Image Understanding}

\begin{document}

\thispagestyle{empty}

\begin{frontmatter}

\title{Towards social pattern characterization in egocentric photo-streams}

\author[1]{Maedeh Aghaei \corref{cor1}}
\cortext[cor1]{Corresponding author: 
  Tel.: +34-672-684-565;}
\ead{aghaei.maya@gmail.com}
\author[1,2]{Mariella Dimiccoli}
\author[3]{Cristian Canton Ferrer}
\author[1,2]{Petia Radeva}
\address[1]{University of Barcelona, Department of Mathematics and Computer Science, Barcelona 08007, Spain}
\address[2]{Computer Vision Center, Bellaterra (Cerdanyola) Barcelona 08193, Spain}
\address[3]{Microsoft
 Research, Redmond, Washington 98052, United States}

\begin{abstract}

Following the increasingly popular trend of social interaction analysis in egocentric vision, this manuscript presents a comprehensive study for automatic social pattern characterization of a wearable photo-camera user, by relying on the visual analysis of egocentric photo-streams. The proposed framework consists of three major steps. The first step is to detect social interactions of the user where the impact of several social signals on the task is explored. The detected social events are inspected in the second step for categorization into different social meetings. These two steps act at event-level where each potential social event is modeled as a multi-dimensional time-series, whose dimensions correspond to a set of relevant features for each task, and LSTM is employed to classify the time-series. The last step of the framework is to characterize social patterns, which is essentially to infer the diversity and frequency of the social relations of the user through discovery of recurrences of the same people across the whole set of social events of the user. Experimental evaluation over a dataset acquired by 9 users demonstrates promising results on the task of social pattern characterization from egocentric photo-streams.

\end{abstract}


\end{frontmatter}

\section{Introduction}


\begin{figure*}[!t]
\centering
\includegraphics[width=\textwidth]{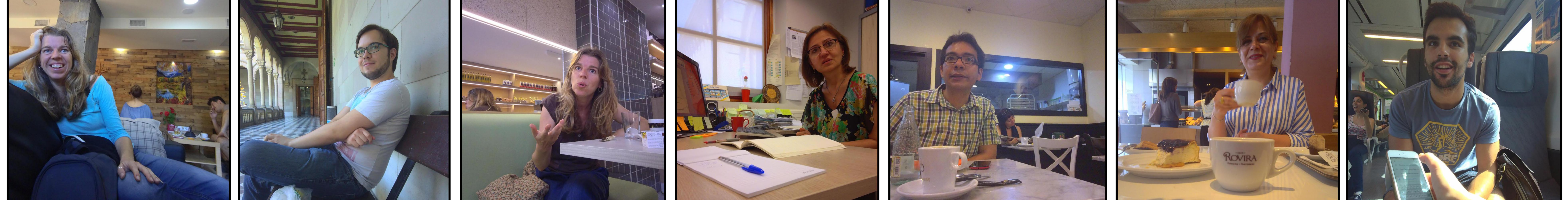}
\caption{Examples of images representing social interactions in our dataset, EgoSocialStyle, captured by the Narrative Clip wearable camera.}
\label{fig:social-interaction}
\vspace*{-3mm}
\end{figure*}

The automatic analysis of  data collected by wearable cameras has drawn the attention of researchers in different topics in computer vision, ranging from object detection and recognition to event summarization and analysis in first-person vision (\cite{betancourt2015evolution, bolanos2017toward}). Among all these topics, social interaction analysis in particular has been an active topic of study (\cite{narayan2014action,alletto2015understanding,aghaei2015towards,aghaei2016whom,fathi2012social,yang2016wearable}). The motivation behind this interest is twofold. Firstly, wearable cameras in comparison to the fixed surveillance cameras, allow to capture natural photos of the daily interactions of the users, where the users naturally attempt to reach a clear view of whom they are engaged in a social interaction (see Fig. \ref{fig:social-interaction}). Hence, the first-person paradigm offers the unique opportunity of revisiting the problem of social interaction analysis from the unmediated first-person view. Secondly, given the strong emotional impact of social interactions, their analysis have a large potential for enabling novel applications in different fields, ranging from entertainment to  preventive medicine. For instance, in a particular scene recorded by a wearable camera, the presence of social interactions is considered as an important factor to determine whether the event is likely to be viewed as worth keeping (\cite{yang2016wearable}). 

The crucial role of personalized characterization of social pattern of a user has been recognized in the medical domain. 
Related works thoroughly investigate the feasibility of using a wearable camera for personalized health monitoring that leads to increase the number of positive clinical outcomes. In this line, \cite{aung2017sensing} and \cite{chow2016sad}  pinpoint how mobile technologies through continuous monitoring, allow  precise assessments of human behavior and ultimately individual mental health. In the same path, \cite{hodges2011sensecam} and \cite{berry2007use} suggested to use wearable cameras for detecting relapse in people affected by depression and \cite{granholm2013ecological} proposed it for ecological momentary assessment of social functioning in schizophrenia. 
Also, in the context of memory training of people affected by mild cognitive impairment, pictures of social interactions are specially treated to trigger autobiographical memory  (\cite{woodberry2015use}). Recently, \cite{dhand2016accuracy} used wearable cameras  for monitoring the lifestyle of stroke survivors and \cite{brown2017snapshot} in a recent study discussed the advantages and disadvantages of incorporating wearable cameras into social psychological research and report data variation on different social situations. In all the aforementioned studies, the key components is to track social interactions of the user in terms of duration and frequency and to monitor their possible variation over time. Indeed, in the literature the importance of \textit{duration} and \textit{frequency} of social interactions in the study of social patterns is well recognized (\cite{carstensen1992social,berry1996positive}). 
Another important factor in social pattern characterization is the study of \textit{diversity} of social interactions which highlights the density of participation of individuals in each kind of formal or informal meetings (\cite{mangrum2001informal,muncy2001disconnecting,steinlin2005knowledge,hudson2006distributing}). Statistical analysis of the social interactions diversity has been considered as a helpful tool to optimize workspace (\cite{steinlin2005knowledge}), to minimize the cost of meetings (\cite{peter2012applying}), and, to maximize effectiveness of interactions among of group members and in the social structure of a broader organization (\cite{oh2004group}).
However, these studies are carried on in non-automatic manners by visual reviewing of the images and other involved signals of interest such as sound. In this work, we introduced a pipeline for automatic analysis of duration, type, frequency and diversity of social interactions in the context of social pattern characterization from egocentric photo-streams.

In sociology, the introduction of the F-formation theory by \cite{kendon1976f} was a foot-stone to formalizing social interaction settings. F-formation is defined as a geometrical pattern that interacting people tend to follow by adjusting their location and orientation towards each other in the space to avoid mutual occlusion. The computer vision community later adopted the F-formation theory to detect groups of interacting people from images and videos (\cite{cristani2011social,gan2013temporal}). Early works about social interaction analysis in conventional images were motivated mainly by video surveillance applications (\cite{setti2015f,cristani2013human}). Surveillance cameras however, capture the environment from a fixed and external perspective and fail in capturing real involvement in social interactions at personal level. Meanwhile, wearable cameras offer the possibility of capturing social cues from a more intimate perspective, known as ego-vision or first-person vision.
Nonetheless, social interaction analysis in ego-vision introduces new challenges in social signal processing in comparison to conventional third-person vision. Unpredictable motion of the camera leads to background clutter and abrupt lighting transitions. In addition, when the frame rate of the camera is low (2 fpm in our case),  drastic visual changes in even temporally adjacent photos make people tracking and their interaction analysis harder (\cite{aghaei2015towards,aghaei2016whom}).

\begin{figure*}[!t]
\centering
\includegraphics [width=\textwidth]{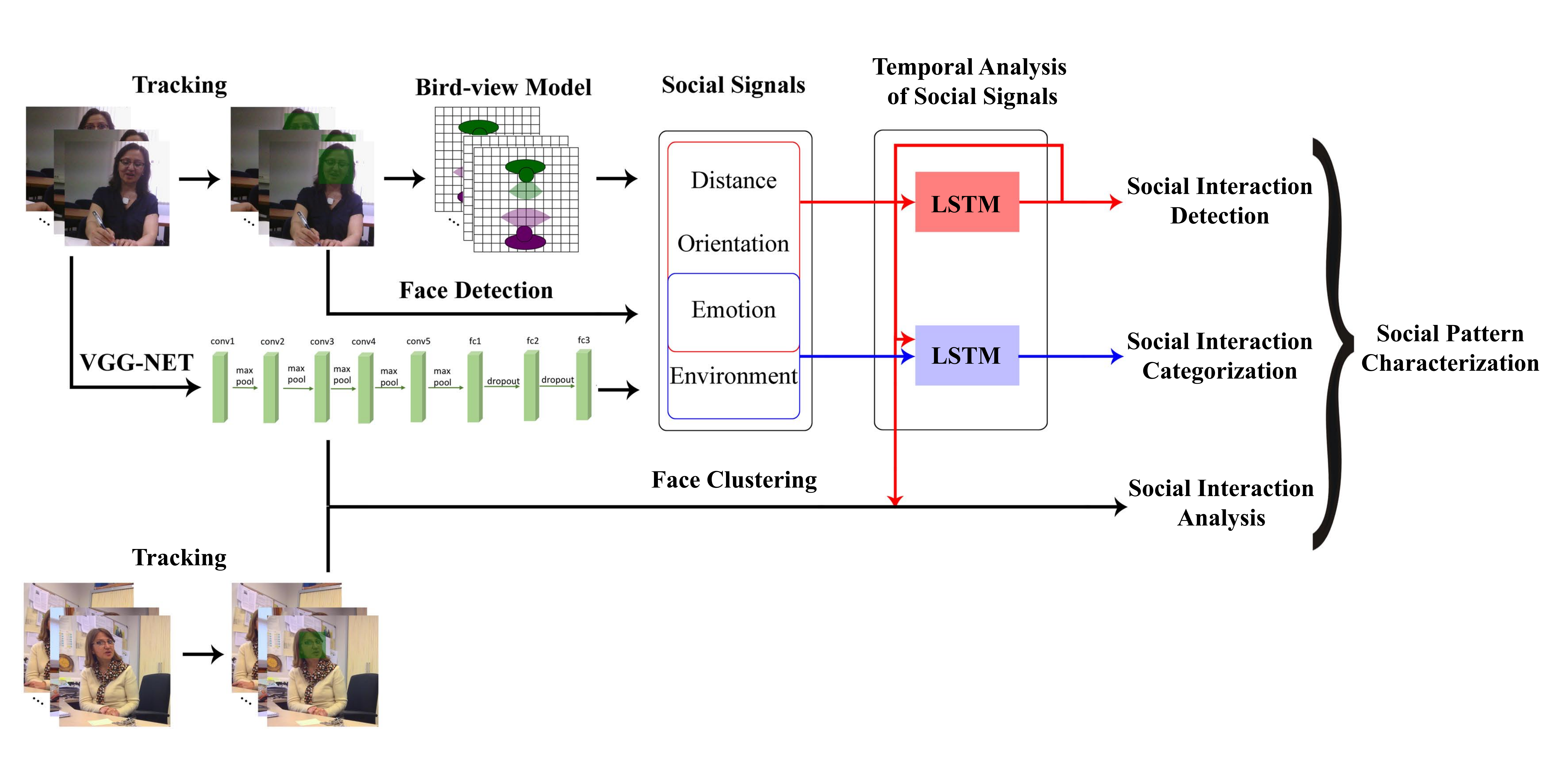}
\caption{Complete pipeline of the  proposed method. Social signals as well as environmental features are extracted for each frame and used to represent each sequence as a time-series. A LSTM is employed to classify each time-series, accordingly to the task at hand: social interaction detection or categorization. Face clustering on the other side enables determination of the diversity and the frequency of social interactions. Eventually, social pattern characterization requires the integration of all tasks.}
\label{fig:pipleline}
\end{figure*}

Building upon our previous work (\cite{aghaei2016whom}), in this paper we go beyond social interaction detection in egocentric photo-streams. The proposed pipeline suggests firstly, to study a wider set of features for social interaction detection and secondly, to categorize the detected social interactions into two broad categories of \textit{meetings}. We focus our attention on the meetings and its two broad \textit{formal} and \textit{informal} subcategories, following the proposed idea by \cite{xiong2005meeting} on social interaction categorization. Eventually, social pattern characterization of the user is achieved through discovery of recurring people in the dataset, and, quantifying the frequency, the diversity and the type of the occurred social interactions with different people. In this work, we prove that to characterize social interactions, analysis of combination of environmental features and social signals transmitted by the visible people in the scene, as well as their evolution over time is required.  A visual overview of the proposed pipeline is given in Fig. \ref{fig:pipleline}. Ideally, employing the entire proposed pipeline in this work, we aim to be able to answer questions such as  \textit{How often does the user engage in social interactions?} \textit{With whom does the user interact most often?} \textit{Are the interactions with this person mostly formal or informal?} \textit{With how many people does the user interact during a month?} \textit{How often does the user see a specific person?} 

Social pattern characterization of individuals requires long term observation of their social interactions, and since wearable photo cameras allow long term recording of the life of a user, they are specifically suitable for this purpose. To demonstrate the generalization ability of the proposed approach, we employ our proposed model over a test set acquired by one user who wore the camera under free-living conditions over one month period while did not participate in acquiring the training set used for training the models. The contributions of this paper can be summarized as follows:


\begin{itemize}
\item Social interaction detection through event-level analysis of different combination of social signals to prove the impact of each signal in the process. The role of facial expression is studied for the first time.
\item Social interaction categorization into formal or informal meetings, considering a set of high-level image features considered as relevant according to an extensive body of sociological literature.
\item Social pattern characterization through the definition and formalization of the frequency, the duration and the diversity of social interactions.
\item Public release of an extensively-annotated egocentric dataset captured in a real-world setting consisting of 125,000 images acquired by 9 users.

\end{itemize}

The rest of the paper is organized as follows: In the next section we provide an overview on the related works to this topic of research, Sec. \ref{sec:detection} is devoted to social interaction detection. Sec. \ref{sec:categorization} details the proposed approach for social interaction categorization and Sec. \ref{sec:beyond} is dedicated to the social pattern characterization. Details about the dataset and experimental results are discussed in Sec. \ref{experiments}. Sec. \ref{egogroup} covers the experimental results over an external dataset, EGO-GROUP, and, Sec. \ref{conclusions} highlights the main conclusions and discusses the future work.

\section{Related work}
\label{sec:related}
The importance of automatic analysis of visual data for the purposes of detection and categorization of social interactions has been recognized by the computer vision community within several studies. Most of the previous studies in social interaction computing were focused on finding potential groups of interacting people, also known as Free-standing Conversational Groups (FCG) in conventional still images or videos. In this regard, \cite{groh2010detecting} proposed to use the relative distance and shoulder orientations between each pair of people to measure social interactions on small temporal and spatial scales. This has been done through training a probabilistic classifier which can then be used for characterizing the social context. \cite{cristani2011social} proposed to solve the task using a Hough-Voting F-Formation (HVFF) strategy to find the common area of interaction by accumulating the density of the overlapping votes of each interacting person. Built upon a multi-scale Hough-Voting policy, \cite{setti2013multi} modeled small FCG as well as large groups of people, relying on different voting sessions.
The problem of finding F-formations has also been formulated as finding dominant sets and using proxemics by employing the graph clustering algorithm (\cite{hung2011detecting}), graph-cuts framework for clustering individuals (\cite{setti2015f}), heat-map based feature representation of interacting people (\cite{gan2013temporal}), and defining an intermediate representation of how people interact (\cite{choi2014discovering}).

The boom of interest in ego-vision during the past few years (\cite{bolanos2017toward}), naturally led to exploration of social interaction analysis in this setting as well. For social interaction analysis in an egocentric scenario, the most exploited features are the  face location and the pattern of attention of the visible individuals, as well as the head movements of the first-person when the camera is worn on the head. \cite{fathi2012social}, proposed a Markov Random Field model to infer the 3D location to which a person is looking at during a social interaction, that relies on the camera intrinsic parameters. They further used this information to classify social interactions into three classes, namely \textit{discussion}, \textit{dialogue} and \textit{monologue}, depending on the active role played by the participants in the interaction. To the best of our knowledge, this is the only previously introduced work about egocentric social interaction categorization. 
Later, \cite{alletto2015understanding} proposed a method for identifying multiple social groups from egocentric videos, that do not rely on the camera intrinsic parameters for 3D projection; hence, the method is  applicable to any head-mounted wearable camera. \cite{soo2015social} introduced the concept of \textit{social saliency} defined as the likelihood of joint attention from a spatial distribution of social members. A social formation is modeled as an electric dipole moment allowing to encode a spatial distribution of social members using a social  formation feature. Recently, \cite{yonetani2016recognizing} proposed to model the dynamics of micro-actions and reactions between two camera-wearer engaged in a dyadic interaction to reach a deeper understanding of the ongoing social interaction between them. In this work, the authors demonstrate that the integration of the first-person perspective of both parties in a dyadic interaction fosters micro-action recognition task in this setting. In another recent attempt,  \cite{yang2016wearable} offered to analyze social interaction sequences and detect them applying a Hidden Markov - Support Vector Machine (HM-SVM). Their focus was on modeling what they called  \textit{interaction features}, mainly physical information of head and body. 

All the aforementioned works share three main common characteristics. First, the high temporal resolution of videos (30-60 fps), which allows to rely on the temporal coherence among video frames to robustly estimate head pose of appearing people and modeling the foreground. Second, the head-mounted cameras, which permits the modeling of head movements and attention patterns of the user. And third, the pursued goal by them, that is restricted to finding potential social \textit{groups} of people in the scene, with exception of \cite{fathi2012social}, that goes deeper into the categorization of social interactions, but strongly relies on head motion for that. Although the high temporal resolution cameras are suitable for capturing details of the action units over relatively short periods of time (up to several hours), they become unfeasible when it comes to social pattern characterization task where long term observation of individuals is required. In this case, low temporal resolution cameras are preferred.


The problem of social interaction analysis from egocentric photo-streams although enables to overcome the aforementioned limitation, due to its particular application has received much less attention (\cite{aghaei2015towards,aghaei2016whom,aghaei2017all}). This problem also introduces novel challenges to the task. Photo-cameras are typically used to acquire photo-streams for long periods of time, thus, are commonly worn on the chest to seek less public attention. Consequently, important information about the head movement of the user is not available and attention estimation becomes unfeasible. In addition, in this particular setting, adjacent frames can present abrupt variations and introduce more difficulty along information processing. In the first attempt towards social interaction detection in egocentric photo-streams, \cite{aghaei2015towards} adapted the HVFF method to the egocentric setting, namely ego-HVFF, to predict social interactions among individuals with the user at frame-level. This method inherently analyzes the social interactions in every frame of the photo-streams separately, and eventually measures the probability of social interaction of the user with each individual based on the ratio of the frames that the algorithm found them as interacting. Later in another attempt to detect social interactions of the user, the authors \cite{aghaei2016whom} proposed to model the temporal coherence of the social signals at sequence-level, by employing a special type of  Recurrent Neural Networks (RNN) known as Long-Short Term Memory (LSTM). According to the F-formation notion, the studied social signals in both of these works are distance and orientation of the individuals with regards to the user. The authors reported that analysis of social signals at sequence-level leads to a better social interaction prediction accuracy.

In this work, we propose a complete pipeline for social pattern characterization of a wearable photo-camera user, where for the first time the role of \textit{facial expressions}, in combination with other conventional social signals is studied  in social interaction analysis. The proposed model relies on the long term observation of social interactions of the user, where multiple social signals aggregate together to achieve a robust social interaction analysis. To the best of our knowledge, this work can be considered as the first comprehensive  social pattern characterization study from a first-person perspective.

\section{Social interaction detection}
\label{sec:detection}

\begin{figure*}[!t]
  \centering
    \subfloat[Social interaction]{\includegraphics[width=\textwidth]{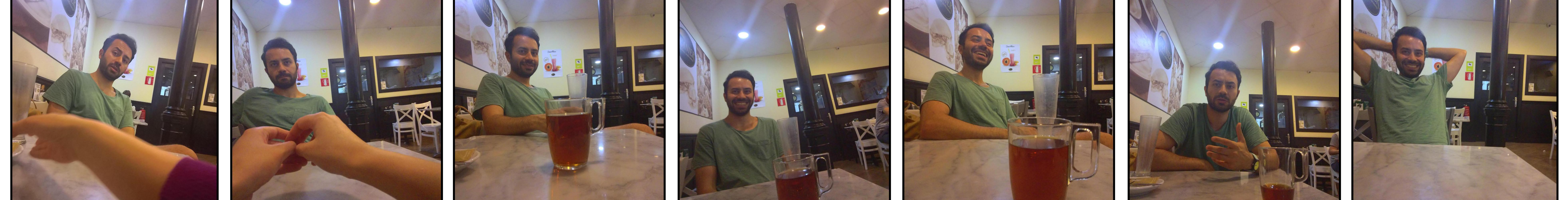}\label{subfig:social}}
    \hfill
    \subfloat[No social interaction]{\includegraphics[width=\textwidth]{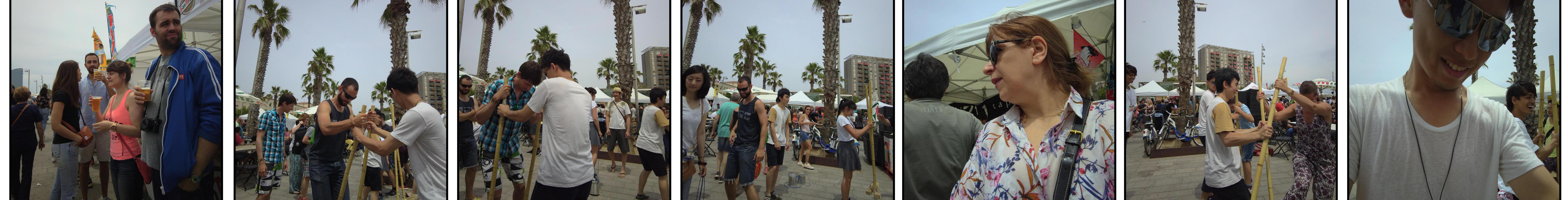}\label{subfig:no-social}}
  \caption{Examples of two sub-sampled sequences in EgoSocialStyle test set. In \ref{subfig:social} the user is involved in a social interaction while \ref{subfig:no-social} demonstrates a sequence where although the user is among the crowd, he is not specifically interacting.}
  \label{fig:social-interaction-vs-not}
\end{figure*}


We, as humans are naturally able to recognize if two or more people are interacting even only by looking at a sequences of images (see Fig. \ref{fig:social-interaction-vs-not}). However,  this is not as trivial for a computer program. In this work, for social interaction detection, we build upon our previous work by introducing additional features and studying their effectiveness in improvement of the results. Specifically, given a \textit{sequence}, a potential social segment of a photo-stream extracted by applying the video segmentation method of \cite{dimiccoli2016sr}, social signals are first extracted at frame-level, and later their evolution is analyzed over time at sequence-level to detect social interactions.



\subsection{Social signal extraction at frame-level}
\label{subsec:ssextraction}

Tracking the appearance of people along time is generally considered as the first step prior to any social behavior analysis in machine vision. In this work, for tracking we employed the extended-Bag-of-Tracklets (eBoT) \cite{aghaei2016multi} which is a multi-person tracking algorithm in egocentric photo-stream setting. The set of bounding boxes corresponding to the same face in a sequence, resulting from eBoT, is called a \textit{prototype}, where the number of prototypes in a sequence is equal to the number of tracked people in it as more than one individual  may appear in a single sequence.

In this work, as well as our previous work, we rely on the F-formation formalization for social interaction detection in the domain of egocentric photo-streams. As the F-formation model assumes a bird-view of the scene, we represent each bounding box in a prototype by a $(x, d, o)$ triplet, so that $x$ denotes the position of the person in the horizontal axis of the image and with regards to the user, $d$ denotes its distance, and $o$ its head orientation. The tracking process, directly provides us with the $x$ position of a face. However, in our egocentric setting, $x$ is not a reliable feature to be considered as it constantly goes under large variations due to the unpredictable movements of the camera and its low frame rate (see Fig. \ref{subfig:social}). Moreover, when it comes to interaction with the user, the $x$ position of the visible people as far as they do not occlude each other, does not play a crucial role. Therefore, we only consider the $(d, o)$ pair to analyze the F-formation. Both parameters, $d$ and $o$  should be calculated for all the participants in the social interaction, being the user and the visible people in a sequence.

\noindent \textbf{Distance:} In the egocentric setting, the user  is obviously located at no distance from the camera $O$ and the distance of the $j$-th tracked person, $p_j$, in the scene from the camera, $d(O,p_j)$, is estimated based on the camera-pinhole model through learning its relation with the vertical face height of the person (\cite{alletto2015understanding}). According to our observations, the relation between the face height of individuals and their distance from the camera is best modeled as a second degree polynomial of the face height of the person  (\cite{aghaei2015towards}).

For training the polynomial regression function, we used the height of the face of 3 different individuals measured in all the following set of  distances $\left \{30, 50, 70, 100, 150, 200, 250  \right \} cm$. The distance feature is represented by: $$\varphi_d(p_j)=d(O,p_j) \in \mathcal{R}.$$
Without loss of generality, in the feature vector we will omit the reference to the person $p_j$ and the wearable camera $O$.

\noindent \textbf{Orientation:} The head orientation of each individual gives a rough estimation of where the person is looking at.  In this work, in addition to the commonly studied yaw ($\omega_z$) head orientation for social interaction detection, pitch ($\omega_y$) and roll ($\omega_x$) head orientations are also studied. Hence, the orientation feature is given by: $$\varphi_o(p_j)=(\omega_x(p_j),\omega_y(p_j),\omega_z(p_j)) \in \mathcal{R}^3,$$ where each of $\omega_x$, $\omega_y$, and $\omega_z$ has a value between [-90\degree,90\degree].
As the camera is basically worn on the chest of the user, we only assume the user can possibly look at anywhere in the space, but with higher probability of looking at other engaged people in  the interaction.

\begin{figure*}[!t]
  \centering
  \subfloat[Social interaction]{\includegraphics[width=0.3\textwidth]{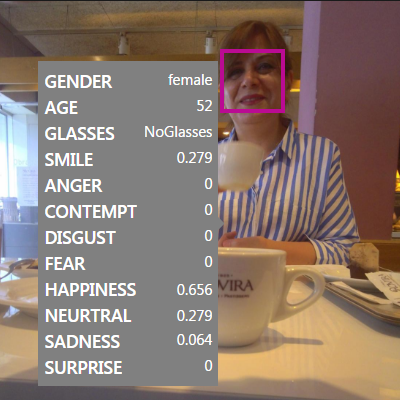}\label{subfig:intemo}}
  \hspace{10mm}
  \subfloat[No social interaction]{\includegraphics[width=0.3\textwidth]{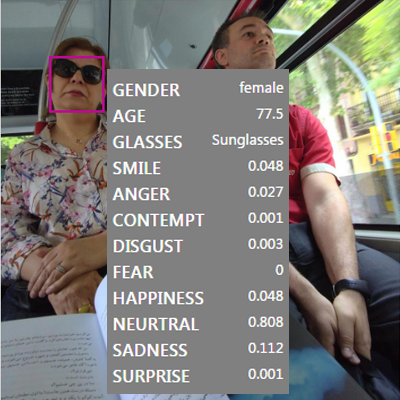}\label{subfig:nointemo}}
  \caption{A same person is shown in two different social events where facial expression probabilities of the person are also presented. When the person is not interacting with the user (\ref{subfig:nointemo}), her dominant facial expression is \textit{Neutral}, while when interacting (\ref{subfig:intemo}) her dominant facial expression varies to \textit{Happiness}.}
  \label{fig:interaction-emotion}
\end{figure*}

\noindent \textbf{Facial expression:} During a social interaction, people exhibit a large number of non-verbal communication cues including facial expressions. Facial expressions as stated by \cite{hess2010you}, are often referred to as automatic demonstrations of affective internal states used as communicative means in interaction with others. The overlooked importance of facial expressions for social interaction detection is mostly noticed within the scenes recorded in crowded places where people often stand in close proximity to strangers with whom they do not necessarily interact. In this situation, relying solely on distance and orientation of the individuals for social interaction detection may lead to disputable predictions (see Fig. \ref{fig:interaction-emotion}). Our observation on real social situations led us to intuitively explore the role of facial expression in social interaction detection as an additional feature beside the pure geometrical features imposed by the F-formation.

In this work, facial expressions and face orientation are extracted by making use of Microsoft Cognitive Service\footnote{https://azure.microsoft.com/en-us/services/cognitive-services/face/}. Facial expression is presented as a predicted vector of probabilities for each of 8 different facial expressions consistently associated to emotions in the occidental culture, being \textit{neutral}, \textit{happiness}, \textit{surprise}, \textit{sadness}, \textit{anger}, \textit{disgust}, \textit{fear},  and \textit{contempt} (\cite{BarsoumICME2016}). For a given person $p_j$, we proposed to consider the index of the dominant facial expression that is a discrete value between 1 (\textit{neutral}) and 8 (\textit{contempt}):
$$ \displaystyle \varphi_e(p_j)= \arg\max_{k \in 1,...,8} e_k(p_j).$$

\subsection{Temporal representation of social signals}
\label{subsec:temporalanalysis}

In this work, the problem of social interaction detection is formulated as a binary time-series classification, where the time-series dimension corresponds to the number of selected social signals for the analysis as explained in Sec. \ref{subsec:ssextraction}. As the complete setting, a $5-$dimensional time-series representing the time-evolution of the $k$-th interaction feature, over time is extracted for each prototype. The task is to classify each time-series as interacting with the user or not. All the aforementioned interaction features, are extracted in every frame of the sequence at time step $\tau$ to build the time-series representation of a prototype: 
$$\varphi_{detection}(\tau,p_j) =(\varphi_d(\tau,p_j),\varphi_o(\tau,p_j),\varphi_e(\tau,p_j)) \in \mathcal{R}^5, \; \tau=1,2,\ldots.$$

\subsection{Time-series classification by LSTM}

Time-series classification is a predictive modeling problem and what makes this problem difficult is that the original sequences can vary in length, be comprised of a very large vocabulary of input symbols and may require the model to learn the long-term context or dependencies between symbols in the input time-series. In this context, RNNs with LSTMs showed great promise to learn the information hidden among steps of a sequence (\cite{ma2016learning}, \cite{jia2015guiding}). LSTM owes its ability to its incorporated memory cells that use logistic and linear units with multiplicative interactions with input and output gates. In this way, it overcomes the exponential error decay problem of RNN and increasing complexity of HMM for learning long term dependencies.

For egocentric sequence binary classification purpose, in this paper we propose to train a LSTM network by introducing to it the time-series from each sequence as presented in previous subsection at each time step. All the aforementioned features for each sequence are introduced to the network as input. The system must learn to classify sequences of different lengths to interacting or not by analyzing the feature vectors associated to each sequence. Hence, the system needs to learn to protect memory cell contents against even minor internal state drift.

\section{Social interaction categorization}
\label{sec:categorization}

\begin{figure*}[!t]
  \centering
    \subfloat[Formal meeting]{\includegraphics[width=\textwidth]{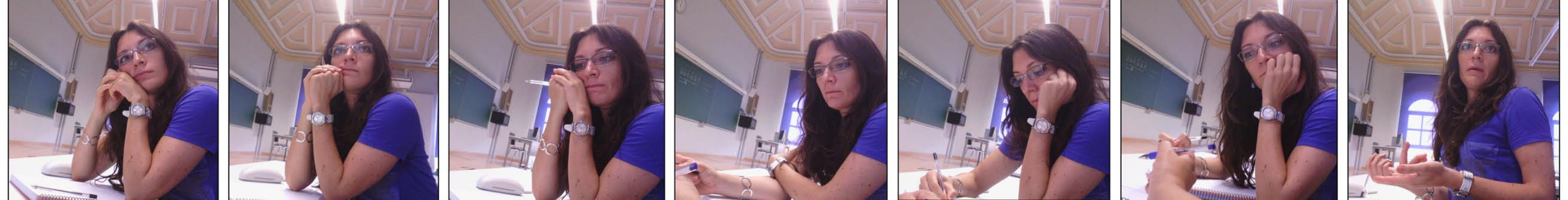}\label{subfig:formal}}
    \hfill
    \subfloat[Informal meeting]{\includegraphics[width=\textwidth]{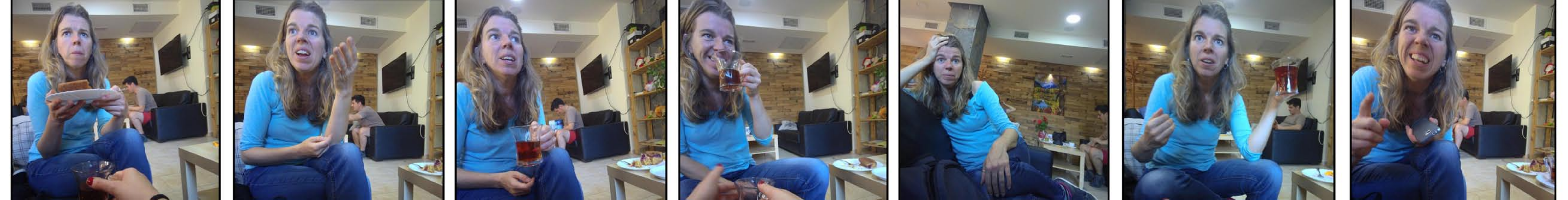}\label{subfig:informal}}
  \caption{Example of two sub-sampled sequences, demonstrating the engagement of the user in different categories of social interactions; a formal meeting (\ref{subfig:formal}), and an informal meeting (\ref{subfig:informal}). The variations in the environment as well as facial expressions of the person in different events can be appreciated.} 
  \label{fig:formal-vs-informal}
\end{figure*}

Social interaction categorization is the task of characterizing type of a social interaction. In the literature, three major elements have been typically exploited for social interaction categorization: the physical setting or place, the social environment, and the activities surrounding the interaction (\cite{palispis2007introduction}). In this work, following \cite{xiong2005meeting} we propose to categorize the detected social interactions in the previous step into two broad categories of common social interactions as \textit{formal meetings} and \textit{informal meetings}, also known as \textit{informal gatherings}. 

\textit{Meetings} are defined as gatherings at which humans communicate, convince, cajole, conspire, and collaborate (\cite{xiong2005meeting}). In general sociology, a formal meeting is defined as a pre-planned event where two or more people come together at a pre-planned place at a particular time to discuss specific matters for the purposes of achieving a specific goal (\cite{xiong2005meeting}). Meanwhile, an informal meeting is more casual, requires less planning, and, usually can take place at any casual space from a park to a hall. Looking closely from the computer vision perspective at the definition of each meeting, environmental features show sign of discriminative power. Therefore, for social interaction categorization we base our approach on the use of environmental features. In addition, we also attempt to study the impact of the facial expressions of involved individuals in the interaction on defining the category of a social interaction. Our  approach  takes into account the temporal evolution of both environmental and facial expression features by modeling them as multi-dimensional time-series, and relies on the classification power of LSTM for binary classification of each time-series into either a formal or an informal meeting.

\subsection{Feature extraction}
\label{subsec:featurextraction}

\noindent \textbf{Global features:} As explained earlier in this section, the surrounding environment of an interaction is considered among the main indicators for categorizing a meeting. Among different features for image representation, CNN features showed exceptional results for global representation of the context in images (\cite{girshick2014rich}). In this work, we represent each image with a feature vector extracted by taking the output of the last fully connected layer of  the VGGNet (VGG16) (\cite{simonyan2014very})  pre-trained  on the Imagenet dataset (\cite{deng2009imagenet}).
However, since the image feature vector consists of thousands of variables, the computational cost becomes significant when it comes to further processing. In addition, the Hughes phenomenon (\cite{hughes1968mean}) is inevitable when it comes to learn a high-dimensional feature space with limited number of training samples in machine learning in general and in RNNs, specifically (\cite{pascanu2013difficulty}). 

In this work, to resolve the curse of dimensionality of CNN features we propose first to apply quantization and then to apply PCA to keep the most important components of the quantization result. To quantize the CNN features, we propose to re-write them as discrete words  as proposed by \cite{amato2016large}. This method takes advantage of the inverted-index approach to deal with the sparsity of the CNN features to associate each component of the feature vector with a unique alphanumeric keyword. This conversion leads to a sparser textual representation of the CNN features in which the relative term  is proportionally related to the feature intensity. This method showed great promises in retrieval applications. CNN feature to word conversion essentially represents each component of the L2-normalized CNN feature vector, $f_k, k = 1, ..., 4096$, as a word: 
$$w_k = {\lfloor} Qf_k {\rfloor},$$
where ${\lfloor} {\rfloor}$ denotes the floor function, and $Q$ is an integer positive quantification factor being $Q>1$. For instance, if we fix $Q = 2$, for $f_k < 0.5$, then $w_k = 0$, while for $f_k\geq 0.5$, $w_k = 1$. The factor $Q$ has a regulator effect on the features for further processing. The smaller the $Q$ the sparser is the new feature vector and it represents less details about the original feature vector. In this work, $Q = 15$ is used which results in highly sparse feature vector representation of integer values: $(w_1, w_2, \ldots, w_{4096}).$

Given the high sparsity of the obtained word representation, a PCA is applied over the so obtained feature vectors extracted from all the images of the dataset and from the emerging representation, 95\% of the most important information are kept. This process results in a 35-dimensional feature vector, $\varphi_{CNN} \in \mathcal{R}^{35}$, while keeping the most important environmental features of the image. Note that applying PCA on the raw CNN features without conversion to word representation, does not result to a feature vector dimension smaller than hundreds. We are interested in keeping the dimensionality of features in the order of tens.

\begin{figure*}[!t]
  \centering
  \subfloat[Formal]{\includegraphics[width=0.5\textwidth]{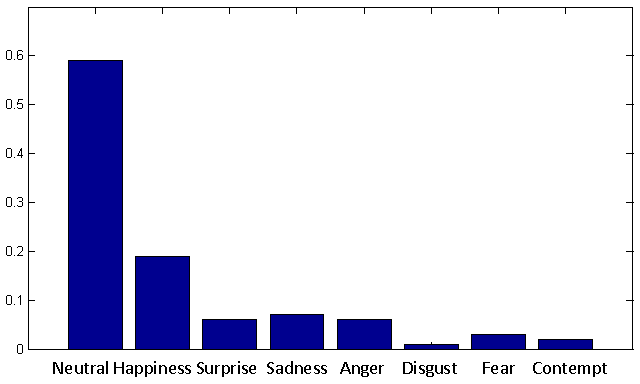}\label{fig:f1}}
  \hfill
  \subfloat[Informal]{\includegraphics[width=0.5\textwidth]{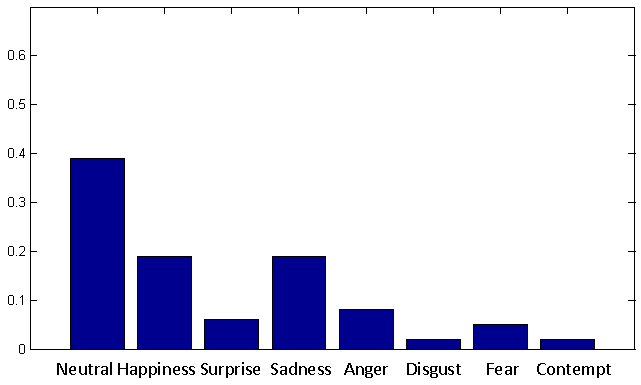}\label{fig:f2}}
  \caption{Bar-plot of facial expression variations over 10 randomly selected sequences for each of (a) formal and (b) informal meetings from the training set in EgoSocialStyle. Each sub-figure shows the mean of the observed facial expressions for each detected face in all the frames of 10 randomly selected sequences. Within informal meetings, people seem to express more freely their emotions as more variation can be observed.}
  \label{fig:formalinformal}
\end{figure*}

\noindent \textbf{Facial expression:} Following our hypothesis that formal and informal meetings can be characterized by the environmental characteristic as well as the facial expression of participants, integration of both features is required. A proof for this hypothesis is illustrated on Fig. \ref{fig:formalinformal} that shows the bar-plot of eight facial expressions for both formal and informal meetings. These bar-plots, obtained using ground truth information, suggest that people express more freely their emotions in informal meetings. Facial expression features in this task are extracted as the mean of facial expressions of the total number of $J$ people detected in each frame of a sequence: 
$$\displaystyle \varphi_{e,k}= \frac{1}{J} \sum_{j = 1}^J e_{k}(p_j), k = 1, \ldots, 8.$$

\subsection{Temporal analysis of representative features}

To achieve joint effect of global image features representing the environment and facial expression features of individuals on social interaction categorization, the 8-dimensional vector of facial expression probabilities ($\varphi_e(\tau)$) is directly concatenated to the environmental features represented by global image characteristics of the event ($\varphi_{CNN}(\tau)$). Given a sequence, the time-series of interaction sequences are constructed as follows for the social interaction categorization:
$$
\varphi_{categorization}(\tau)  =(\varphi_{CNN}(\tau), \varphi_e(\tau)) \in \mathcal{R}^{43}, \; \tau=1,2,\ldots
$$

Further, time-series classification task into either a formal or an informal meeting is reached relying on the LSTM power for time-series classification.

\section{Social pattern characterization}
\label{sec:beyond}

\subsection{Generic social interaction characterization}
\label{sec:beyond_generic}

Characterizing the social pattern  of an individual, implies the ability of defining the nature of social interactions of the user from various temporal (frequency, duration, etc.)  and social (type, identity, and, number of interaction people with the user, etc.) aspects.  Providing a definition within the aforementioned contexts, demands social interaction analysis of the user across several events during a long period of time. For this purpose,  we define four concepts to characterize social interactions, namely \textit{frequency}, \textit{social trend}, \textit{diversity}, and \textit{duration}.

\noindent{\bf Frequency (F):} Defines the normalized rate of formal (or informal) interactions of a person by the total number of observation days:
$$F_{formal (informal)}=\#{formal (informal) \; interactions}/\#{days}$$

\noindent{\bf Social trend (A):} Indicates whether the majority of social interactions of a person are formal (or informal):
$$A_{formal (informal)}=\#{formal (informal)\; interactions}/\#{all \; interactions}$$ 

\noindent{\bf Diversity (D):} Demonstrates how diverse are social interactions of a person. The term is defined as the exponential of the Shannon entropy calculated with natural logarithms, namely: 
$$D =1/2 \exp \left(-\sum _{i\in\{formal,informal\}}A_{i}\ln(A_{i})\right)$$
Note that when the person has the same number of formal and informal interactions (i.e. $A_{formal}=A_{informal}=0.5$),  $D=1$.

\noindent{\bf Duration (L):} Defines the longitude of each social interaction of the user. The duration is the longitude of the sequence corresponding to the $i$-th social interaction, say  $L(i) = \mathcal{T}(i)r$, where $\mathcal{T}(i)$ is the number of frames of the $i$-th interaction and $r$ is the frame rate of the camera.
Different statistics can be applied on the duration of interactions like mean, median or  standard deviation in order to characterize social interactions and extract the social pattern.

\subsection{Person-specific social interaction characterization}

\begin{figure*}[!t]
\centering
\includegraphics[width=\textwidth]{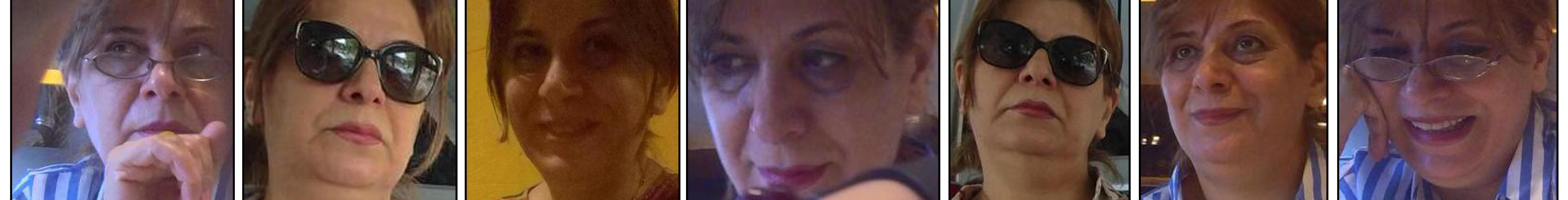}
\caption{A few examples of faces belonging to one cluster obtained by applying \cite{aghaei2017all} on the EgoSocialStyle test set. The visual variation among face examples can be appreciated.}
\label{fig:cluster}
\end{figure*}

In this subsection, we consider the concepts for social interaction characterization of the user within the context of interaction with a specific person. This firstly requires that all the interactions of the user with a certain person to be localized. To this goal, a face clustering method to find various appearances of the same person among all the social events of the user  (\cite{aghaei2017all}) is employed. The face clustering method is applied on the results of the social interaction detection step. To cope with the extreme intra-class variability of faces, it builds upon the multi-face tracking outcome.  In a single event, tracking gathers a set of different appearances of the same face in that event, called a \textit{face-set} in this context, which allows to reshape the face clustering task in different events to face-set clustering.

The deterministic factor in deciding whether two different face-sets belong to the same cluster, i.e. represent the same person, is defined through a dissimilarity measure. Let $R$ and $T$ be a \textit{reference} and a \textit{target} face-sets, respectively. Let $S^R$ be the similarity matrix between all possible pairs of face-examples in $R$, and $S^T$ be the similarity matrix between face-examples in $R$ and face-examples in $T$. The dissimilarity between $T$ and $R$, $\delta(R,T)$,  is calculated as the absolute difference between the median value $\mu$ of $S^R$ and $S^T$, respectively:
$$\label{eq:diss}
 \delta(R,T) = \left | \mu ^R - \mu ^T \right |.
$$

A hierarchical clustering technique is applied to group the face-sets according to their pair-wise dissimilarity value. The cut-off threshold for the agglomerative clustering is chosen empirically over a separate learning dataset and corresponds to the median value of all dissimilarities between the face-sets corresponding to the same person. Fig. \ref{fig:cluster} shows a few images in one resulting cluster obtained together with an index to which sequence each element of the cluster belongs. One can appreciate the visual variance of the faces in a cluster.

\subsection{Face-cluster analysis}
Let  $\mathcal{C} = \left \{c_j\right \}$, ${j = 1, \ldots, J}$ be the set of clusters obtained by applying the face-set clustering method  on the detected interacting prototypes, where $J$ ideally corresponds to the total number of people who appeared in all social events of the user along the whole period of observation (e.g. a month). Each cluster, $c_j$, ideally contains all the different appearances of the person $p_j$ across different social events, and $|c_j|$ is the cardinality of $c_j$ which demonstrates the number of social interactions events of the user with the person $p_j$ during the observation period. 

As both the clustering method and the proposed method for social interaction detection and categorization act at sequence-level, inferring the interaction state of each sequence inside a cluster is straightforward. 
The frequency, the social trend, the diversity and the duration of the interactions with a specific person, can be computed in the same manner as explained in \ref{sec:beyond_generic}, by restricting the interactions considered to the ones with the  person of interest.

\section{Experiments and discussion}
\label{experiments}
In this section, we introduce our dataset for social pattern characterization in egocentric photo-streams, namely EgoSocialStyle and describe the proposed experimental setup to validate our proposed approach. A comprehensive discussion to provide broader insight over the obtained results is also given in this section.

\subsection{Experimental setting}

\subsubsection{Data}
\label{subsec:data}

To the best of our knowledge, this work is the first attempt to characterize automatically the social pattern of a person relying exclusively on visual data. Lack of  previous studies goes with the lack of the public dataset on this considered purpose, which led us to build a new dataset to validate jointly all the tasks of our proposed method. Our dataset has been acquired by 9 users wearing a Narrative clip camera during the participation in gathering the dataset while they were living their daily life without any  constrains. The camera was set to automatically capture a photo every 30 seconds once being worn. The participants who gathered the dataset had different ages and profiles and wore the camera in different and random days and times of the week. Sequences in our dataset have different lengths, varying from 20 to 60 frames (10 to 30 minutes of interactions).

The training set of EgoSocialStyle is an extended version of the dataset previously introduced in \cite{aghaei2016whom}. 
It has been acquired by 8 users; each user wore the camera for a number of non-consecutive days over a total of 100 days period, collecting over 100,000  images in total, where in 3,000 images among them a total number of 62 different persons appear.

The test set is acquired by a single user, who did not participate in acquiring the training set as we aimed to study the generalization ability  of our model for social pattern characterization of a person. The user wore the camera for 30 consecutive days collecting 25,200 images, where  2,639  of which correspond to social events. There are 35 sequences with more than one person appearing in them over 113, in total. 40 different trackable persons appear in the test set. 

Face annotations in the whole dataset are attained using the Microsoft face annotation tool (\cite{BarsoumICME2016}). 
Participants were asked to provide a label (interacting/not interacting, formal/informal) for their own sequences. Table \ref{table:dataset} provides further details of the proposed dataset.

\begin{table*}[!t]
\centering
\caption{EgoSocialStyle dataset consists of train set and test set captured by 9 different users. The details about each set is provided in this table.}
\label{table:dataset}
\begin{tabular}{lccccccccc}
\hline
                                  
\begin{turn}{40}\#\end{turn}                                    &\begin{turn}{40}Users\end{turn}        & \begin{turn}{40}Days \end{turn}           & \begin{turn}{40}Images\end{turn}       & \begin{turn}{40}\begin{tabular}[c]{@{}l@{}}Social\\ Images\end{tabular} \end{turn}           & \begin{turn}{40}People \end{turn}       & \begin{turn}{40}Sequences\end{turn}       & \begin{turn}{40}Prototypes\end{turn}      & \begin{turn}{40}Interacting\end{turn}     & \begin{turn}{40}Formal\end{turn}         \\ \hline
{Train} & {8} & {100} & {100,000} & {3,000} & {62} & {106} & {132} & {102} & {42} \\ 
{Test}  & {1} & {30}  & {25,200}  & {2,639}  & {40} & {113} & {172} & {130} & {25} \\ \hline
\end{tabular}
\end{table*}

\subsubsection{Data augmentation}
Large amount of data for better training of deep models is a well recognized necessity. However, the required time to acquire and label real data for this purpose is not negligible and is where artificial data augmentation could have an impact. A proper data augmentation is one which provides a reasonable set of data in addition and similar to the already existing data in the training set, but also slightly different from them to reduce overfitting of the model in learning a task (\cite{wong2016understanding}). Besides the impact of data augmentation in the production of additional data, it is also considered a helpful tool to provide balance to unbalanced data. This specially is of interest in our case where to acquire sequences without any social interaction is more difficult than sequences with social interaction.


To augment the data at hand, we employed the proposed idea by \cite{krizhevsky2012imagenet}. The principle idea consists of augmenting signals by adding slight variations to them, which can be done by adding eigen-features on top of each different feature in a sequence. This has been achieved through applying PCA and then adding multiples of the found principal components to each sequence, with magnitudes proportional to the corresponding eigenvalues times a random variable drawn from a Gaussian with mean zero and small standard deviation (0.01, in this work). This scheme  generates more data in addition to the original training data by applying label-preserving transformations to them.

Let  $\Phi= (\varphi_{1,n}(\tau),\varphi_{2,n}(\tau),\ldots,\varphi_{K,n}(\tau)),$ 
$n=1, \ldots, N$  is the set of all the $N$ time series in our training set where $\tau=1,\ldots, \mathcal{T}$, is the length of the sequences and consequently the time-series and, $k=1, \ldots, K$, is the dimension of the time-series. Note that in the social interaction detection task, $N$ is equal to the total number of prototypes in the training set, and in the social interaction categorization task, $N$ is equal to the number of sequences in the training set.

The augmentation of $\Phi$ from $N$ to $\hat{N}$, with $\hat{N} = \Delta N$,  is achieved through adding the vector $\hat{\Phi}_n(\tau) = (\phi_{1,n}(\tau),\phi_{2,n,}(\tau),\ldots,\phi_{K,n}(\tau))$ to the  frame $\tau$ of the $n$-th time-series in $\Delta$ number of attempts. $\hat{\Phi}_n(\tau)$ is obtained as: $$\hat{\Phi}_{n}(\tau)=[P_1, P_2, \ldots, P_K][\theta_{1,n}(\tau)\lambda_1,\theta_{2,n}(\tau)\lambda_2,\ldots,\theta_{K,n}(\tau)\lambda_K]^T,$$  where $P_k$ and $\lambda_k$ are the $k$-th eigenvector and eigenvalue of the K × K covariance matrix of feature values, respectively, and $\theta_{k,n}(\tau)$ is the aforementioned random variable. It is worth to mention that in the social interaction detection task, $K = 4$ and in the social interaction categorization task, $K = 32$. In the social interaction detection, since the facial expression is a variable with discrete values, we did not consider to alter it in the data augmentation. Instead, when we generated new samples of time-series from an original time-series, we only repeated the facial expression signal of the original time-series in the augmented time-series. We did not consider to alter the facial expression signal neither in the social interaction categorization task, since the facial expression feature vector originally contains values of probabilities which must sum to 1 and altering them leads to a change in their essence. Instead, similar to the other tasks, we only repeated the facial expression signal of the original time-series in the augmented time-series.

\subsubsection{Network structure and hyper-parameter optimization}

In this work, we used the most commonly used version of LSTM in literature, known as vanilla LSTM (\cite{graves2005framewise}) for time-series classification. This architecture is a three layer network consisting of the input layer, the LSTM hidden layer and a sigmoid output layer, where the input layer has forward connections to all units in the hidden layer and each LSTM is composed of various numbers of memory cells. We added a dropout layer between the hidden layer and the output layer to mitigate the overfitting problem. Vanilla LSTM in contrary to the first introduced version of LSTM (\cite{hochreiter1997long}), features \textit{forget gate} in addition to \textit{input gate} and \textit{output gate}. It also incorporates peephole connections and uses full Backpropagation Through Time (full-BPTT) instead of truncated gradient training. 

In vanilla LSTM, the output of the LSTM block is recurrently connected back to the block input and all of the gates, but it does not use \textit{full gate recurrence} as in the initial version of LSTM. Full gate recurrence means that all the gates receive recurrent inputs from all gates at the previous time-step which greatly increases the number of parameters that has been discouraged in the literature (\cite{greff2017lstm}). Stochastic Gradient Decent method (SGD) is used for optimization in full-BPTT training. As both of the social interaction detection and categorization tasks are binary classification problems, we used an output layer with a single neuron and a sigmoid function to make 0 or 1 predictions and a log loss as the loss function. Due to the higher computational complexity of the gate specific dropout techniques in the hidden layer, we did not use any of them.

Different settings of features require different settings of hyperparameters to give good performance, and we are interested in the best performance that can be achieved with each setting. For this reason we chose to tune the hyperparameters for each setting, separately. Grid search with 3-fold cross validation on the training set has been used in order to obtain best performing hyperparameters. The studied parameters for the grid-search are learning rate, momentum, dropout rate, batch size, number of epochs, and, number of LSTM blocks per hidden layer. We made log-uniform sampling over the following interval of hyper-parameters: [0.0001,0.1] learning rate, [0.1,0.9] momentum, [0.0,0.9] dropout rate, [100,1000] batch size,  [10,100] epochs, and, [10,200] number of LSTM blocks. The best performing hyperparameters per each setting for each task are given in Table \ref{table:hyperparameters}.

\begin{table*}[!t]
\centering
\caption{Best performing hyperparameters for each setting per each task. Social interaction detection settings are separated by a horizontal line from social interaction categorization settings.}
\label{table:hyperparameters}
\begin{tabular}{@{}lcccccc@{}}
\toprule
     & Learning rate & Momentum & Dropout rate & Batch size & Epoch & \#Cells \\ \midrule
SID1 &       0.001        &    0.7      &     0.0         &      20      &    50   &     30   \\
SID2 &       0.01        &      0.8    &       0.0       &      30      &   50    &     35   \\
SID3 &       0.001        &     0.7     &      0.5        &      50      &    100   &    30    \\
SID4 &       0.001        &     0.5     &      0.0        &      20      &    100   &    100    \\ \hline
SIC1 &       0.001        &    0.8      &      0.0        &     50       &    50   &    200    \\
SIC2 &       0.001        &     0.9     &      0.0        &     50       &    20   &    150    \\
SIC3 &       0.01        &     0.8     &       0.5       &      100      &    50   &    200    \\ \bottomrule
\end{tabular}
\end{table*}

\subsection{Experimental results and discussion}
\label{results}

As mentioned earlier, each dimension of a time-series is variation of a unique feature along the sequence. In this section, to prove the importance of each feature and to discover the optimal combination of features, we train and test individual networks by introducing time-series composed of different combination of features.

\subsubsection{Social interaction detection}

In this task, four set of settings are explored as:
\begin{itemize}
\item\textbf{SID1:} Distance + Yaw
\item\textbf{SID2:} Distance + Yaw + Pitch + Roll
\item\textbf{SID3:} Distance + Yaw + Facial expression  
\item\textbf{SID4:} Distance + Yaw + Pitch + Roll + Facial expressions
\end{itemize}

\begin{table*}[!t]
\centering
\caption{Social interaction detection results. The best results in terms of precision, recall and, accuracy are achieved through training and testing the model on the SID4 setting.}
\label{tab:detection} 
\begin{tabular}{@{}lccccc@{}}
\toprule
                                       & ego-HVFF          & SID1         & SID2         & SID3         & SID4         \\ \midrule
{Precision}  & {82.75\%} & {80.76\%} & {88.49\%} & {88.59\%} & {\textbf{91.66\%}} \\  
{Recall}  & {55.81\%} & {64.61\%} & {76.92\%} & {77.69\%} & {\textbf{84.61\%}} \\  
{Accuracy}  & {58.38\%} & {61.62\%} & {75.00\%} & {75.58\%} & {\textbf{82.55\%}} \\ \bottomrule
\end{tabular}
\end{table*}

SID1 is the baseline setting in which only presented features in our previous work (\cite{aghaei2016whom}) are studied.
 In SID2, pitch and roll in addition to yaw as the main indicator of face orientation in previous works are studied.
SID3 follows the same pattern as SID1, but includes facial expression features to observe the effect of facial expressions in addition to commonly studied features for social interaction detection. Finally, SID4 includes all the discussed features for social interaction detection analysis. It is important to note that the data augmentation is only performed once for the complete 4-dimensional setting (SID4) and data in other settings is formed by selecting the required dimensions from the complete setting.

In Table \ref{tab:detection}, we report the obtained precision, recall and accuracy values for each of the above settings. Besides, we also compared our obtained results with the ego-HVFF model as the unique method amongst state-of-the-art methods suitable for social interaction detection in egocentric photo-streams as discussed in Sec. \ref{sec:related}. The best obtained results, in all terms of precision, recall and accuracy belong to the SID4 setting  containing all the proposed features (distance, yaw, pitch, roll, facial expressions) for social interaction detection. Comparing SID1 with each of SID2 and SID3 shows that the incorporation of each of the other head orientation information and facial expression in the analysis leads to more robust social interaction detection, while facial expression shows to have a slightly stronger impact (SID3) than additional head orientations (SID2). Ego-HVFF only considers distance and yaw orientation (SID1) for social interactions detection. However as expected, temporal analysis of SID1 in sequence-level leads to more accurate social interaction detection than  frame-level analysis of the sequences as it has been achieved through applying ego-HVFF. Our reasoning is that since in this task all the social signals originate from the face appearance of the third-person, face occlusions due to movements of the camera or the user itself, lead to social signals discontinuity. Therefore, analysis of the sequences in frame-level results in direct exclusion of occluded frames from the analysis while sequence-level analysis in format of time-series mitigates the social signals fragmentation impact by considering the relation among the rest of the frames of a sequence.

\begin{figure*}[!t]
  \centering
    \subfloat[Correctly detected as no-social interaction employing SID3 and SID4, incorrectly detected as social interaction employing SID1]{\includegraphics[width=\textwidth]{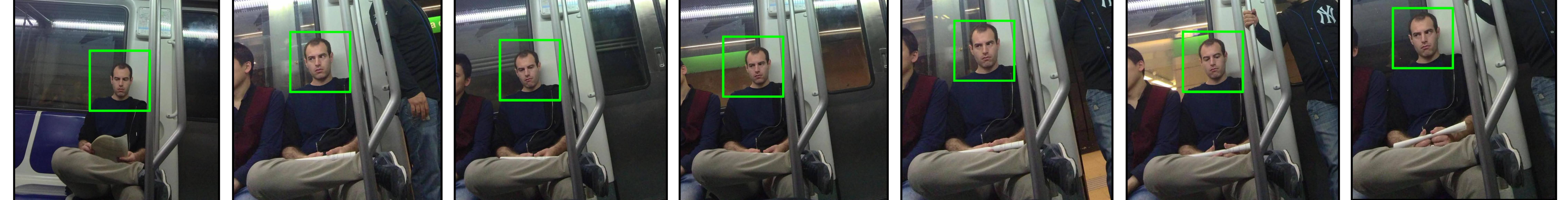}\label{subfig:det-emo-1}}
    \hfill
    \subfloat[Correctly detected as no-social interaction employing SID3 and SID4, incorrectly detected as social interaction employing SID1]{\includegraphics[width=\textwidth]{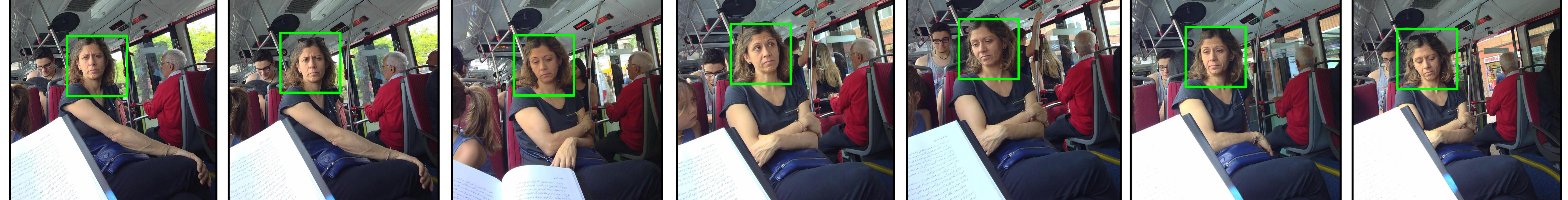}\label{subfig:det-emo-2}}
  \caption{Two examples to highlight the role of facial expression. We assume the invariant \textit{Neutral} facial expression of the individual led to classification success employing both SID3 and SID4 settings, and classification failure employing SID1 setting which does not include facial expression information. For better observability in the cluttered scene, face examples of the individuals are shown by a green bounding boxed around them.} 
  \label{fig:detection-emotion}
\end{figure*}

\begin{figure*}[!t]
  \centering
    \subfloat[Correctly detected as social interaction employing SID2 and SID4]{\includegraphics[width=\textwidth]{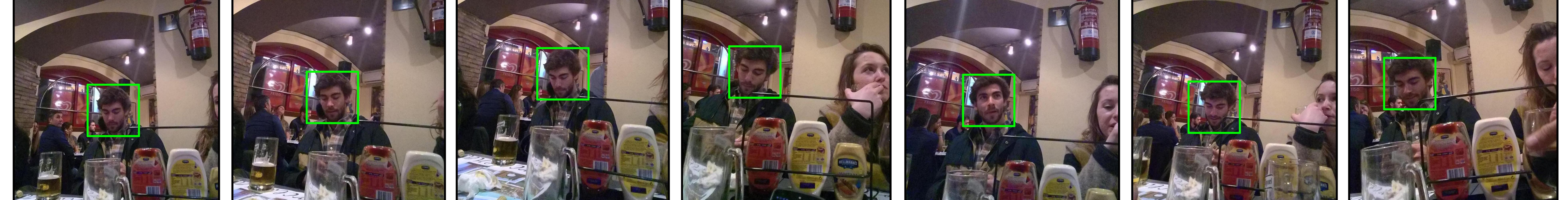}\label{subfig:det-pr-1}}
    \hfill
    \subfloat[Correctly detected as no-social interaction employing SID2 and SID4]{\includegraphics[width=\textwidth]{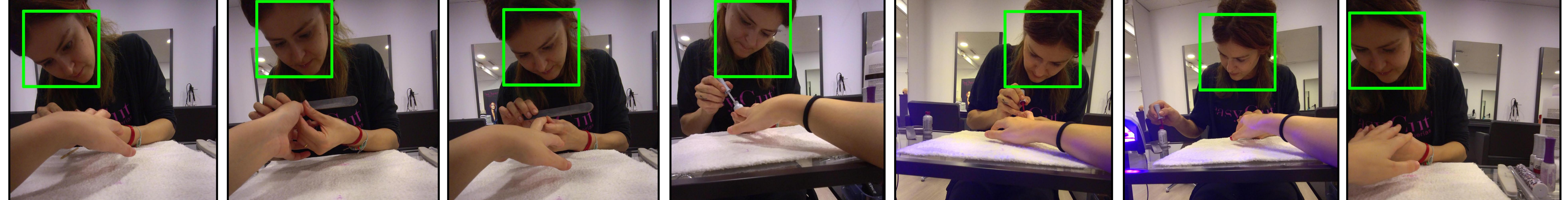}\label{subfig:det-pr-2}}
  \caption{Two examples to emphasize the role of pitch and roll head orientation in social interaction detection. Sequences are  correctly classified employing both SID2 and SID4 settings, and incorrectly classified employing SID1 setting which lacks pitch and roll head orientation information.}
  \label{fig:detection-pr}
\end{figure*}

\begin{figure*}[!t]
  \centering
    \subfloat[Incorrectly detected as no-social interaction employing any of the settings]{\includegraphics[width=\textwidth]{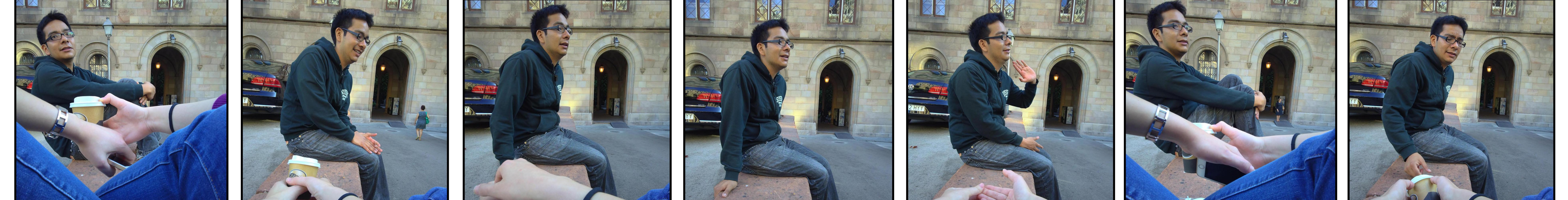}\label{subfig:det-fail-1}}
    \hfill
    \subfloat[Incorrectly detected as no-social interaction employing any of the settings]{\includegraphics[width=\textwidth]{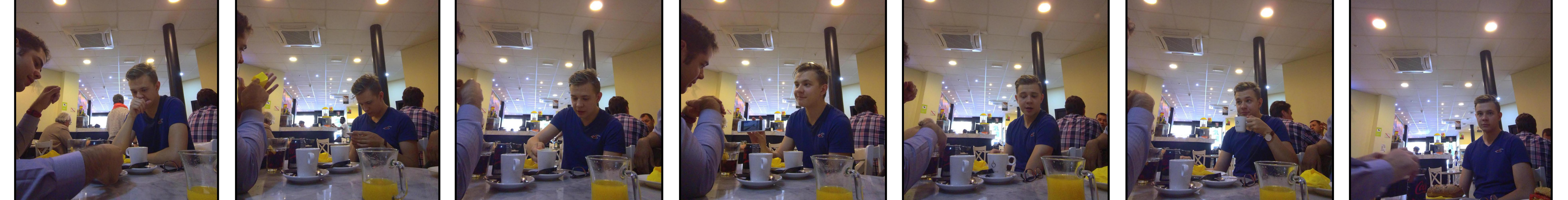}\label{subfig:det-fail-2}}
  \caption{Examples of two sub-sampled sequences in our dataset, where sequences could not be correctly detected as interacting employing any of the settings. The uncommon head pose of the individuals in both sequences led to the model failure.}
  \label{fig:detection-failure}
\end{figure*}

Fig. \ref{fig:detection-emotion} and Fig. \ref{fig:detection-pr} are visual demonstrations of how facial expressions and additional head orientations aid in more robust social interaction detection. In Fig. \ref{subfig:det-emo-1} and Fig. \ref{subfig:det-emo-2}, although the subjects are oriented towards the user and they are in relatively close proximity to the camera, we assume their invariant \textit{neutral} facial expressions were a determinant factor in helping the model to correctly classify them as not interacting with the user. Another scenario can be observed in Fig. \ref{subfig:det-pr-1} and Fig. \ref{subfig:det-pr-2}. In Fig. \ref{subfig:det-pr-2}, despite the close proximity of the subject to the user and although her yaw orientation in inclined towards the user, we assume the uncommon pitch orientation of her head aided the model to correctly classify the sequence as not interacting with the user. Two failure cases of the detection model can be observed in Fig. \ref{fig:detection-failure}. This could happen due to the uncommon head pose of the interacting people and their dominant \textit{neutral} facial expression. Indeed in none of the examples, the interacting people are looking towards the user.

\subsubsection{Social interaction categorization}

\begin{table*}[!t]
\begin{center}
\centering
\caption{Social interaction categorization results. The best results in terms of precision, recall and, accuracy are achieved through training and testing the model on the SIC3 setting.}
\label{tab:categorization} 
\begin{tabular}{@{}lccccc@{}}
\toprule
                                   & HM-SVM    & VGG-FT         & SIC1         & SIC2         & SIC3                   \\ \midrule
{Precision} & {76.82\%} & {86.81\%}  & {87.91\%} & {89.01\%} & {\textbf{91.48\%}}  \\  
{Recall} & {63.65\%} & {89.77\%} & {90.90\%} & {92.04\%} & {\textbf{97.72\%}}  \\  
{Accuracy} & {64.87\%} & {82.30\%} & {83.18\%} & {84.95\%} & {\textbf{91.15\%}} \\ \bottomrule
\end{tabular}
\end{center}
\end{table*}

In this task, the following settings are considered for the temporal analysis:
\begin{itemize}
\item \textbf{SIC1:} Environmental (VGG)
\item \textbf{SIC2:} Environmental (VGG-finetuned)
\item \textbf{SIC3:} Environmental (VGG-finetuned) + Facial expressions
\end{itemize}


We assume that global features of an event, namely environmental features, have the greatest impact in the categorization of it. Therefore in this section, the first setting (SIC1) studies only environmental features which are extracted from the last fully connected layer of the VGGNet trained over the Imagenet and preprocessed as explained in Sec. \ref{subsec:featurextraction}.  VGGNet trained on the Imagenet is highly capable of grasping the general semantics in an image. However, fine-tuning the network for a specific task over relevant data for that task, adapts the pre-trained network to that specific purpose. Therefore, we assume the extracted features from the fine-tuned network ideally lead to better representation of the desired classification task. In SIC2, the environmental features are extracted in the same manner as SIC1, but from the fine-tuned VGGNet over the training set of the EgoSocialStyle. The features are preprocessed in the same manner as explained in Sec. \ref{subsec:featurextraction}. Fine-tuning the network is achieved through instantiation of the convolutional part of the model up to the fully-connected layers and then training fully-connected layers on the photos of the training set. The last setting to be studied is SIC3, which explores jointly the effect of facial expressions as well as the environmental features in social interaction categorization.

In this task, we have employed VGGNet pre-trained over Imagenet for feature extraction, while any other CNN architecture suitable for image feature extraction could be employed and finding the optimal CNN architecture is out of scope of this work. Additionally, the Imagenet dataset was preferred to a seemingly more relevant dataset such as Places (\cite{zhou2014learning}) for environmental feature extraction of images. This is due to the narrow field of view of the Narrative camera where in the images captured by it, a scene is better observed by the set of visible objects in it rather than the wide view of the scene.

In Table \ref{tab:categorization}, we report the precision, recall and accuracy values obtained for each setting of the aforementioned settings. Additionally, we compared our obtained results with HM-SVM (\cite{yang2016wearable}) which is an applicable state-of-the-art method to our setting as this model similarly to ours  extracts features in the egocentric setting and analyzes them in sequence-level but different to our proposed model, employs a HMM to model interaction sequences according to features to categorize them. To apply HM-SVM, the HMM is trained using our training set where features follow the SIC3 setting. The HM-SVM is later employed to label the interaction state. We also report achieved results by a baseline method, VGG-FT, in which the fine-tuned VGG network on the photos of the training set in EgoSocialStyle is tested over the pool of photos in EgoSocialStyle test set. Thus, it is considered a frame-level modeling of the problem.

The obtained results suggest that, temporal analysis of environmental features extracted from fine-tuned VGGNet in SIC2 setting outperforms temporal analysis of environmental features extracted from VGGNet before fine-tuning in the SIC1 setting. Temporal analysis of fine-tuned features also outperforms frame-level analysis of fine-tuned features in VGG-FT which is also an indication of the importance of temporal analysis of features in this task. The combination of environmental features extracted through fine-tuned VGG network and feature vector of facial expressions probabilities leads to the highest performance of the model. HM-SVM is trained and tested with features in the SIC3 setting. However, the obtained results suggest that the LSTM demonstrates more power in modeling the problem at hand than the HMM.

It is worth to note that, due to the extensive amount of data that end-to-end models need for training (few million data) and to our limited number of image sequences in the dataset, we did not consider to design our proposed model in an end-to-end fashion. Indeed, making use of pre-trained networks, like emotion, makes a more effective use of the resources when the available data is small compared with the amount of data needed to train the individual sub-networks.

\begin{figure*}[!t]
  \centering
    \subfloat[Correctly detected as informal meeting employing SIC3]{\includegraphics[width=\textwidth]{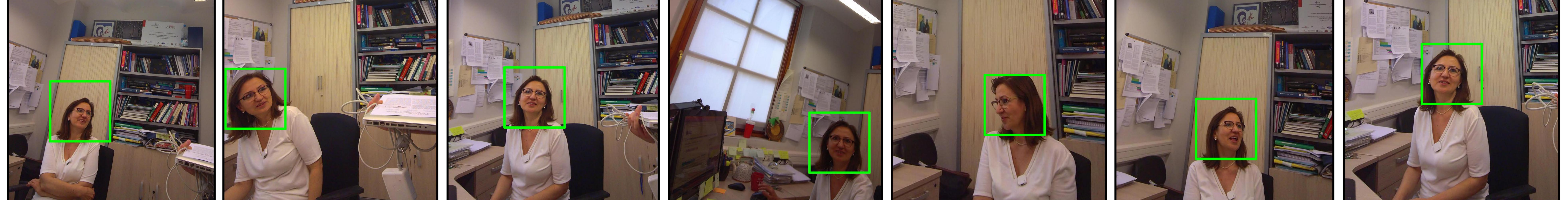}\label{subfig:cat-1}}
    \hfill
    \subfloat[Correctly detected as formal meeting employing SIC3]{\includegraphics[width=\textwidth]{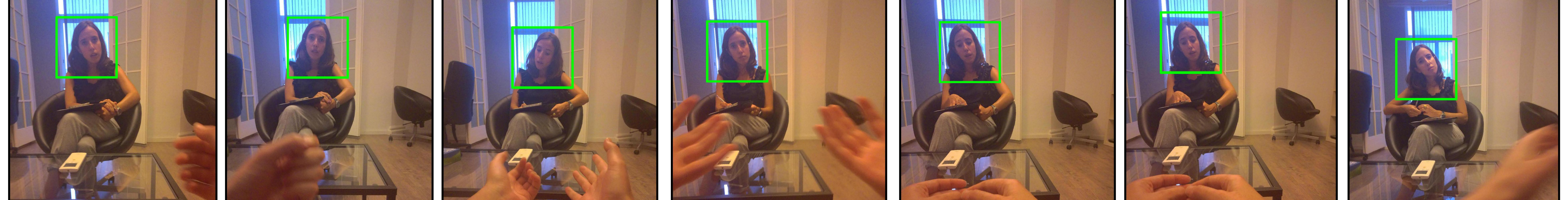}\label{subfig:cat-2}}
  \caption{Two successful examples employing SIC3 setting, emphasizing on the role of facial expressions in social interaction categorizations. The method trained over mere general features employing SIC2 setting did not lead to the right categorization of each of the sequences.} 
  \label{fig:cat-results}
\end{figure*}

\begin{figure*}[!t]
  \centering
    \subfloat[Incorrectly detected as formal meeting employing SIC3]{\includegraphics[width=\textwidth]{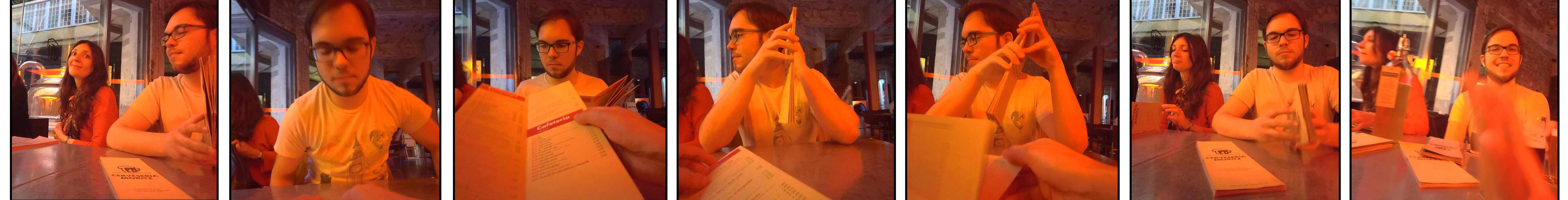}\label{subfig:failure-1}}
    \hfill
    \subfloat[Incorrectly detected as formal meeting employing SIC3]{\includegraphics[width=\textwidth]{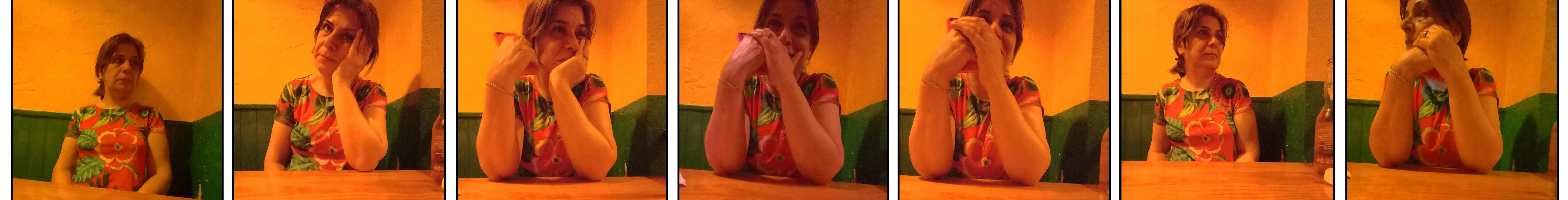}\label{subfig:failure-2}}
  \caption{Two failure examples of the model trained on any of the social interaction categorizations settings. We assume misleading environmental features in \ref{subfig:failure-1} and invariant \textit{neutral} facial expressions of the subject in \ref{subfig:failure-2} led to these failure cases.} 
  \label{fig:failure}
\end{figure*}

In Fig. \ref{fig:cat-results}, two sequences are shown in which the aggregation of facial expressions with the general environmental features employing SIC3 leads to the correct categorization of them. In Fig. \ref{subfig:cat-1}, although the environment is the indicator of a formal meeting, we assume the variant facial expressions of the subject aids the model to correctly classify it as an informal meeting. On the contrary, in Fig. \ref{subfig:cat-2} despite the scene not implying a formal meeting, we assume the dominant \textit{neutral} facial expression of the subject leads to the correct categorization of the sequence as a formal meeting.  Fig. \ref{fig:failure} shows two cases where the model fails to correctly categorize  social interactions due to  misleading  features  transmitted from the scene. Both Fig. \ref{subfig:failure-1} and Fig. \ref{subfig:failure-2} are informal gatherings which are classified incorrectly as formal meetings. Our assumption is that in Fig. \ref{subfig:failure-1} the model confuses the menu with a piece of paper which is an important characteristic of a formal meeting. We also assume in Fig. \ref{subfig:failure-2} the invariant \textit{neutral} facial expression of the person leads  the model to fail.

\subsubsection{Social pattern characterization}

\begin{figure*}[t]
\centering
\includegraphics[width=0.7\textwidth]{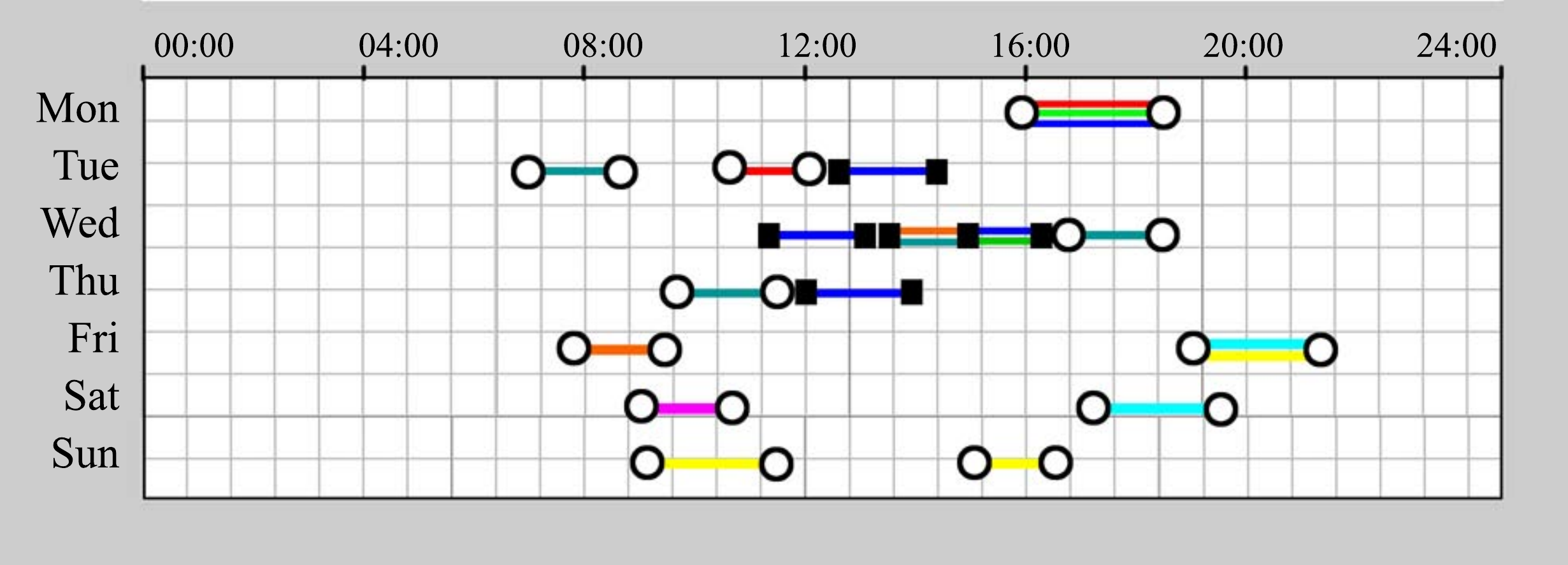}
\caption{Temporal map of social interactions of the user during one week. The boundaries of an interaction are shown by circles for informal and squares for formal interactions. Different line colors are index of the interaction with different people and multiple lines within a boundary are indicative of interaction with multiple people.
}
\label{fig:characterization}
\end{figure*}

To illustrate the ability of the proposed framework for social pattern characterization of an individual, face clustering is applied on the test set. A total number of 83 clusters is obtained, which is almost double the size of the total number of prototypes in the test set. The largest cluster contains 77 number of faces from 5 number of sequences belonging to the same person in various social events. 

The different statistics of the social interactions of the user, as well as those related to the most frequently interacted person are  provided in Table \ref{table:characterization}. From our observation, it can be concluded that during the observation interval the user interacted with the most frequently interacting person 5 times, in 4 different days, 4 times of which  occurred during informal meetings. An interesting observation is that in a cluster containing different sequences, a sequence may belong to a formal or informal meeting which implies the user may have different types of interaction with the same person in various social events. According to the statistics reported in Table \ref{table:characterization}, generic diversity of social interaction of the user is relatively high (87\%). Specifically, the user  is three times more inclined towards having informal meetings than formal meetings (Generic A-Formal vs. A-Informal, 0.75 vs. 0.25) and thus, more frequently gets engaged in informal meetings as supported by the statistics. Interestingly, the generic social trend of the user is correlated to the person-specific one (0.05 difference in both formal and informal social trends). The above interpretation is expected when assuming an informal social interaction can occur at any time without any planning, while for formal social interactions normally planning is involved (\cite{xiong2005meeting}). 

The social pattern of the user over one week according to the obtained results from clustering and inference to their types is visualized in Fig. \ref{fig:characterization}. Social interactions are shown by horizontal colored lines, where the interaction boundaries are shown by circles for informal meetings and squares for formal meetings. Different colors correspond to different persons. Re-occurring people in one social event are shown with parallel lines within the same interval. As it can be observed in Fig. \ref{fig:characterization}, informal social interactions of the user are happening at almost any time of the day and the formal social interactions are normally happening during the middle of the day.

\begin{table*}[!t]
\centering
\caption{Social pattern characterization results, demonstrating the generic and person-specific frequency (F), social trend (A), diversity (D) and, Duration (L) of the social interactions of the user.}
\begin{tabular}{@{}lcccccc@{}}
\toprule
                & F-Formal & F-Informal & A-Formal & A-Informal & D & L     \\ \midrule
Generic         & 0.83             & 2.50               & 0.25                & 0.75                  & 0.87      & 25.19±1.32   \\
Person-specific & 0.25             & 1.00               & 0.20                & 0.80                  & 0.59      & 18.80 ± 0.96 \\ \bottomrule
\end{tabular}
\label{table:characterization}
\end{table*}

\section{Social pattern characterization on EGO-GROUP}
\label{egogroup}

Despite the lack of available datasets for the purpose of social pattern characterization in egocentric vision, to demonstrate the effectiveness of our proposed model, we applied the entire pipeline on EGO-GROUP (\cite{alletto2015understanding}), a most adaptable public dataset to our considered purpose in this work. Despite the fact that EGO-GROUP is not designed dataset for computing the statistics of social style of a user (social pattern characterization), it offers a benchmark that is directly suitable for social interaction detection and adaptable for social interaction categorization in the domain of egocentric vision. 

\begin{figure*}[!t]
\centering
\includegraphics[width=\textwidth]{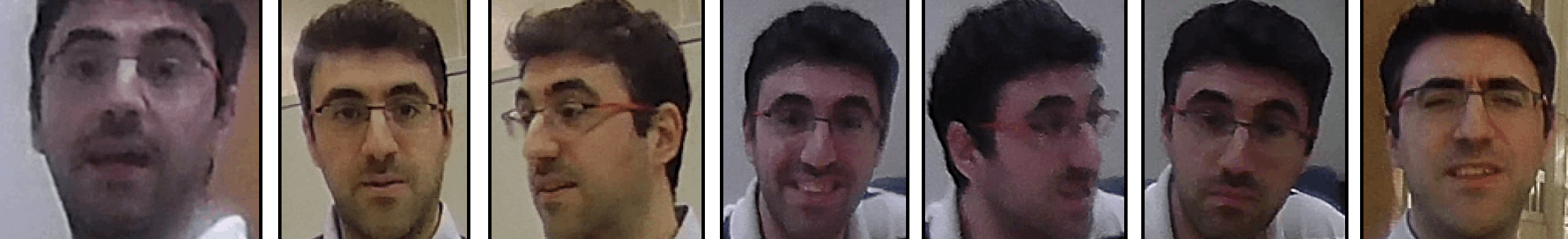}
\caption{A few examples of faces belonging to the biggest cluster obtained by applying \cite{aghaei2017all} on the EGO-GROUP dataset. Face-examples in this clusters belong to three different scenarios of EGO-GROUP.}
\label{fig:group-cluster}
\end{figure*}

EGO-GROUP is a social group detection dataset for egocentric vision, which consists of 18 videos collected in five different scenarios: laboratory, coffee break, conference room, outdoor, and party. The ground truth data available with the dataset in addition to the type of each scenario, provides interaction labels for each individual. To adapt the dataset to the definition of social interaction category in this work, we labeled the laboratory and the conference room videos as formal meeting, and, party, coffee break, and, outdoor as informal meeting scenarios. However, as mentioned before, social pattern characterization purpose requires long term monitoring of daily life of a person. Whereas, EGO-GROUP consists of single detached by scenario sequences that are captured under controlled, and not free living conditions. For this reason, in this section we report the obtained results for social interaction detection and categorization as well as face clustering.

For the sake of a fair comparison, we down-sample the videos captured in 15 fps to 1 fps photo-streams. Within the terminology used in this paper, we obtained 21 social events (sequences) and 76 prototypes. For social interaction detection, we followed the same proposal in the Sec. \ref{sec:detection}, with the only difference that the distance feature is calculated as it is proposed in the original paper (\cite{alletto2015understanding}). For social pattern categorization, we used one event of each scenario for fine-tuning the network and used the new fine-tuned network to extract the word representation of training set for training the LSTM. Later, the appropriately trained LSTM is used for testing the model. For both of social interaction detection and categorization tasks, we only evaluate the models on the best performing settings of features on EgoSocialStyle, being SID4 in the case of social interaction detection and SIC3 in the case of social interaction categorization. We report the obtained results on EGO-GROUP in terms of precision, recall, and, accuracy in Table \ref{table:ego-group}. 

EGO-GROUP does not provide any clustering ground truth to validate this task. However, as part of the entire framework we also applied the clustering on this dataset. Examples of the face-examples in the biggest obtained cluster are shown in Fig. \ref{fig:group-cluster}. This cluster contains 86 face-examples of the same person from several events across three different scenarios in EGO-GROUP.

\begin{table}[]
\centering
\caption{The obtained results in terms of precision, recall, and, accuracy on the best performing settings for both tasks of social interaction detection (SID4) and categorization (SIC3) on EGO-GROUP. }
\label{table:ego-group}
\begin{tabular}{@{}lcc@{}}
\toprule
          & Detection  & Categorization  \\ \midrule
Precision & 86.11\% & 90.00\% \\
Recall    & 77.50\% & 75.00\% \\
Accuracy  & 81.57\% & 76.47\% \\ \bottomrule
\end{tabular}
\end{table}


\section{Conclusions}
\label{conclusions}

In this work, we proposed a complete pipeline for social pattern characterization of a user wearing a wearable camera for a long period of time (e.g. a month), 
relying on the visual features transmitted from the captured photo-streams. 
Social pattern characterization is achieved through first, the detection of social interactions of the user and second, their categorization. In the end, different appearances of interacting individuals with the user are localized across different social events  through face clustering to directly drive the frequency and the diversity of social interactions of the user with each individual. In the proposed method, social signals for each task are presented in the format of multi-dimensional time-series and LSTM is employed for  the social interaction detection and categorization tasks. A quantitative study over different combination of features for each task is provided, unveiling the impact of each feature on that task.  Evaluation results suggest that in comparison to the frame-level analysis of the social events, sequence-level analysis employing LSTM leads to a higher performance of the model in both tasks. 

To the best of our knowledge, this is the first attempt at a comprehensive and unified analysis of social pattern of an individual in either ego-vision or third-person vision. This comprehensive study can have important applications in the field of preventive medicine, for example in studying social patterns of patients affected by depression, of elderly people and of trauma survivors.

\section*{Acknowledgments}
This work was partially funded by TIN2015-66951-C2-1R, SGR 1219, Grant 20141510 (Marat\'{o} TV3), and CERCA Programme / Generalitat de Catalunya. M. Aghaei is supported by APIF grant of University of Barcelona. M.Dimiccoli and P. Radeva are partially supported by ICREA Academia 2014. The authors acknowledge NVIDIA Corporation for the donation of GPUs. The funders had no role in the study design, data collection, analysis, and preparation of the manuscript.

\bibliographystyle{model2-names}
\bibliography{bibliography}

\end{document}